\documentclass[letterpaper, 10 pt, journal, twoside]{IEEEtran}
\usepackage{amsmath,amssymb,amsfonts}
\usepackage{eucal}
\usepackage{pgfplots}
\pgfplotsset{compat=1.18} 
\usepackage{graphicx} 
\usepackage{textcomp}
\usepackage{xcolor}
\usepackage[hidelinks]{hyperref}
\usepackage{authblk}
\usepackage{verbatim}
\usepackage{algorithm}
\usepackage[noend]{algpseudocode}
\usepackage{multirow}
\usepackage{booktabs}
\usepackage{tabularx}
\usepackage{gensymb}
\usepackage{lipsum}

\usepackage{algpseudocode}
\usepackage{soul}
\usepackage{mathptmx, mathtools}
\usepackage{subcaption}
\usepackage[utf8]{inputenc}
\usepackage[nodisplayskipstretch]{setspace}
\usepackage{cite}
\usepackage{multicol}
\usepackage{multirow}
\usepackage{subcaption}
\usepackage{caption}
\usepackage{listings}
\usepackage{tcolorbox}
\usepackage{placeins} 

\usepackage[font=small,skip=1.0pt]{caption} 
\usepackage{array}

\newcommand\subparagraph{%
  \@startsection{subparagraph}{5}
  {\parindent}
  {3.25ex \@plus 1ex \@minus .2ex}
  {-1em}
  {\normalfont\normalsize\bfseries}}
\makeatother
\usepackage{titlesec}
\let\subparagraph\relax 
\titlespacing\section{0pt}{4pt plus 1pt minus 1pt}{0pt plus 1pt minus 1pt}
\titlespacing\subsection{0pt}{4pt plus 1pt minus 1pt}{0pt plus 1pt minus 1pt}
\titlespacing\subsubsection{0pt}{4pt plus 1pt minus 1pt}{0pt plus 1pt minus 1pt}

\def\BibTeX{{\rm B\kern-.05em{\sc i\kern-.025em b}\kern-.08em
    T\kern-.1667em\lower.7ex\hbox{E}\kern-.125emX}}

\begin{document}

\title{Robotic Compliant Object Prying Using Diffusion Policy Guided by Vision and Force Observations}


\author{
Jeon Ho Kang$^{1}$, Sagar Joshi$^{1}$, Ruopeng Huang$^{1}$, and Satyandra K. Gupta$^{1}$%
\thanks{\textcopyright2025 IEEE. Personal use of this material is permitted. Permission from
IEEE must be obtained for all other uses, in any current or future media,
including reprinting/republishing this material for advertising or promotional
purposes, creating new collective works, for resale or redistribution to servers or lists, or reuse of any copyrighted component of this work in other works.}
\thanks{Manuscript received: November 19, 2024; Revised January 31, 2025; Accepted March 2, 2025.}
\thanks{This paper was recommended for publication by Editor Pascal Vasseur upon evaluation of the Associate Editor and Reviewers' comments.}%
\thanks{$^{1}$ J.H. Kang, S. Joshi, R. Huang, and S.K. Gupta are with the Viterbi School of Engineering, University of Southern California, Los Angeles, USA.
{\tt\footnotesize \{jeonhoka, guptask\}@usc.edu}}%
\thanks{Digital Object Identifier (DOI): see top of this page.}%
}

\markboth{IEEE Robotics and Automation Letters. Preprint Version. Accepted March, 2025}
{Kang \MakeLowercase{\textit{et al.}}: Robotic Compliant Object Prying Using Diffusion
Policy Guided by Vision and Force Observations} 

\maketitle

\begin{abstract}
The growing adoption of batteries in the electric vehicle industry and various consumer products has created an urgent need for effective recycling solutions. These products often contain a mix of compliant and rigid components, making robotic disassembly a critical step toward achieving scalable recycling processes. Diffusion policy has emerged as a promising approach for learning low-level skills in robotics. To effectively apply diffusion policy to contact-rich tasks, incorporating force as feedback is essential. In this paper, we apply diffusion policy with vision and force in a compliant object prying task. However, when combining low-dimensional contact force with high-dimensional image, the force information may be diluted. To address this issue, we propose a method that effectively integrates force with image data for diffusion policy observations. We validate our approach on a battery prying task that demands high precision and multi-step execution. Our model achieves a 96\% success rate in diverse scenarios, marking a 57\% improvement over the vision-only baseline. Our method also demonstrates zero-shot transfer capability to handle unseen objects and battery types. Supplementary videos and implementation codes are available on our project website: \url{https://rros-lab.github.io/diffusion-with-force.github.io/}\\

\end{abstract}

\begin{IEEEkeywords}
Deep Learning in Grasping and Manipulation, Sensor Fusion, Disassembly
\end{IEEEkeywords}

\section{Introduction}

\noindent \IEEEPARstart{I}t is becoming increasingly critical to refurbish, reuse, and recycle products. In order to maximize material recovery and minimize environmental impact, an effective method for disassembly is essential. However, disassembly poses more challenges than assembly due to unknown assembly states. Moreover, disassembly operations require significant troubleshooting due to issues such as worn, rusted, or corroded parts. When humans perform these tasks, they rely on multiple sensing modalities, such as vision and force.

\begin{figure}
    \centering
    \noindent\includegraphics[width=\linewidth]{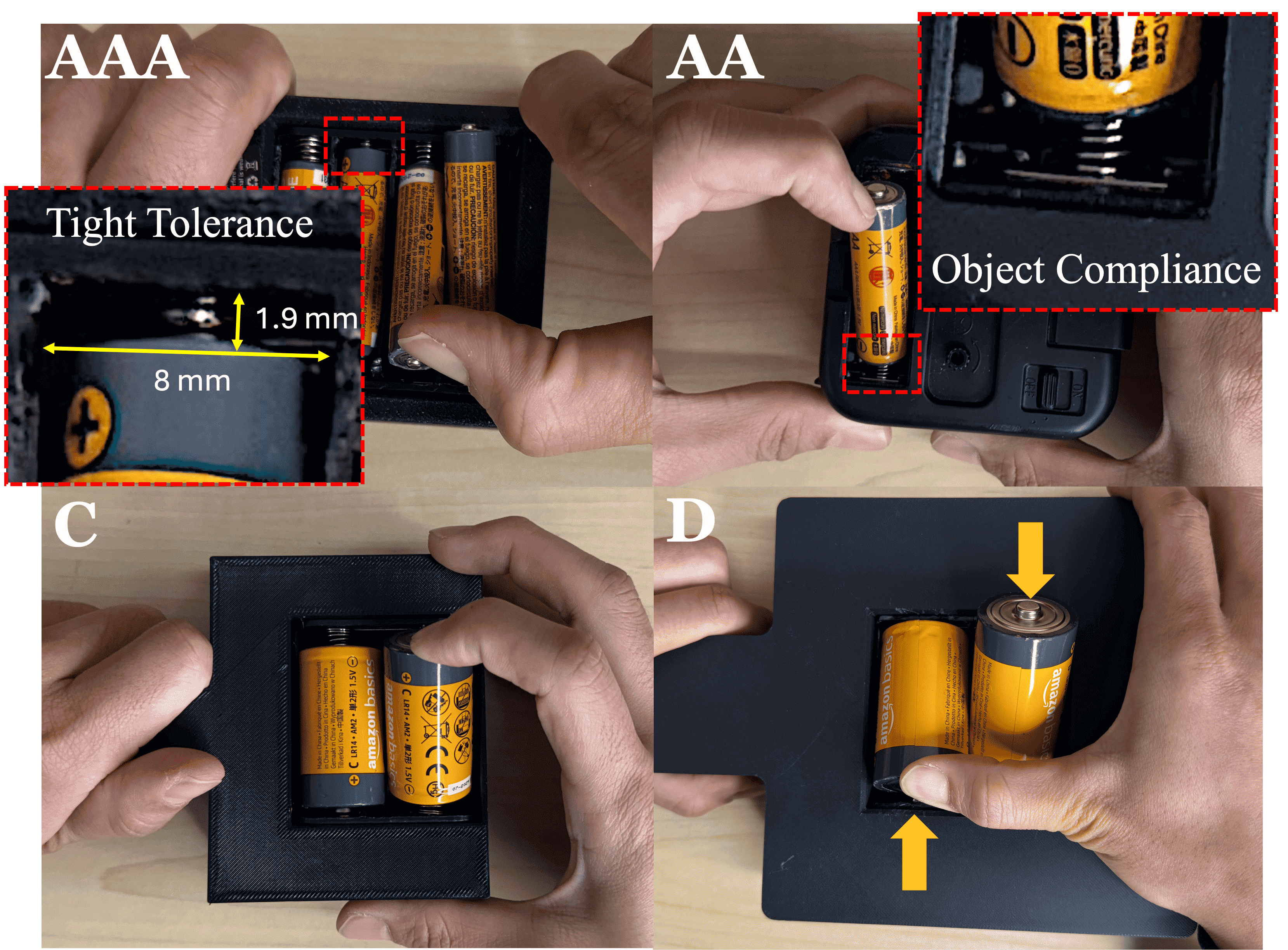}

    \caption{Humans use both arms to perform compliant object prying. In this figure, a human demonstrates battery prying for four different battery types, highlighting the tight tolerances and high precision required. A spring on one end introduces compliance to the object, and dependence on direction.}
    \label{fig:human_prying}
\end{figure}

Given these complexities, robotic disassembly faces significant challenges. However, still higher level of automation is necessary for scalable disassembly. Traditional offline programming approaches for robotic disassembly fall short because many tasks demand high precision and real-time feedback, similar to human capabilities \cite{fang2019survey}. 

One approach to achieving automation is imitation learning, which involves acquiring skills from human experts. While effective for relatively simple tasks \cite{argall2009survey}, traditional imitation learning faces challenges in scenarios which has multiple ways to achieve the same goal—sometimes referred to as multi-modality. Recent advances show that diffusion models address this issue effectively by committing to a single mode when multiple plans exist \cite{chi2023diffusionpolicy, chi2024diffusionpolicy, chi2024universal, liu2024maniwavlearningrobotmanipulation}.

This paper addresses the force-based compliant object prying task, specifically in battery disassembly, which involves a robot executing a tilting motion while applying adequate contact force to deform a component, enabling it to separate from its assembly. These products are housed in spring-loaded or tightly fastened casings, as illustrated in Fig. \ref{fig:human_prying}. Performing this task presents several challenges due to tight tolerances and variability in the approach, tilt, and insertion of the prying tool. These factors are influenced by the size of the object and the compliance of the assembly, making it difficult for traditional methods to generalize effectively. 

There has been prior work that has studied battery prying problem using a rule-based method \cite{10378967}. This approach uses programmed Cartesian motions with force feedback to perform a prying motion. While effective, it is robust primarily in scenarios where the state information of the battery is precise and tolerances are high. In contrast, we aim to take a learning-based approach that generalizes to diverse scenarios involving multiple types of batteries and configurations. Because our method learns a visuo-motor policy, our policy can adapt more flexibly to varying force requirements associated with different battery types and positional variations. To this end, we propose a learning from demonstration framework for disassembly tasks, enabling adaptability and robustness.

We utilize a diffusion policy with vision and force feedback as learning from demonstration framework, enabling the robot to perform generalizable contact-rich, prying tasks in compliant assembly. When integrating vision and force into diffusion policy, disparity in dimensionality can lead to the lower-dimensional input becoming diluted, reducing its influence. To address this issue, we use cross-attention mechanisms that learn relational features between vision and force signals, creating more expressive observations. Our method demonstrates a significantly improved success rate compared to benchmark approaches, outperforming naive force-based methods and achieving results comparable to human demonstration regarding force application patterns and task execution time. Our contributions include:

\begin{enumerate}
    \item  A novel method using cross-attention architecture to incorporate force into observation space in diffusion policy action prediction
     \item Force signal augmentation techniques to account for variability during inference, improving robustness with out-of-distribution objects
    \item Successful application of diffusion policy to compliant object prying, achieving a 96\% success rate across both seen and unseen objects and battery types
\end{enumerate}

\section{Related Works}
\label{Related Works}
\noindent \textbf{Behavior Cloning}:  Learning from demonstration is a broad approach in robotics where an agent learns from actions done in expert demonstrations \cite{argall2009survey}. One popular approach in learning from demonstration, behavior cloning is a supervised learning approach where an agent learns to map states to actions provided by an expert \cite{bain1995framework}. It has demonstrated remarkable potential across various real-world robotic manipulation tasks \cite{zhang2018deep, mandlekar2020learning, zeng2021transporter, seita2020deep, rahmatizadeh2018vision, florence2019self, schulman2016learning}. Explicit behavior cloning treats learning as regression task, directly aiming to minimize the difference between predicted actions and expert demonstrations \cite{florence2022implicit}. On the other hand, implicit policy models action distributions by assigning energy values to actions, selecting those that minimize energy \cite{florence2022implicit, jarrett2020strictly, du2019implicit}. Behavior Transformer (BET) \cite{shafiullah2022behavior} leverages transformer architecture to model sequential decision-making by capturing long-range dependencies. However, learning multi-modal action distribution has been a challenge in these behavior cloning approaches \cite{mandlekar2020iris, florence2022implicit,shafiullah2022behavior}.

\noindent \textbf{Diffusion Models for Policy Learning}: Diffusion models are probabilistic generative models designed to produce output by gradually transforming noise into data points that match the target distribution \cite{sohl2015deep, ho2020denoising}. \cite{ wang2023diffusion} uses diffusion model to effectively learn policies in offline reinforcement learning, outperforming traditional methods by balancing behavior cloning with policy improvement. \cite{reuss2023goal, pearce2023imitating} implement diffusion models to imitate human behavior in simulated environments. Diffusion models are also used for tasks involving robotic manipulation \cite{urain2023se, black2023zero, kapelyukh2023dall, mishra2024reorientdiff}, and \cite{mishra2023generative, sridhar2024nomad} use diffusion models to solve planning problems, demonstrating the potential of diffusion models in generating efficient plans for complex tasks.  \cite{chi2023diffusionpolicy, chi2024universal} leverage diffusion models for visuo-motor policy learning on physical robots with impressive results compared to other methods. \cite{wu2024tacdiffusionforcedomaindiffusionpolicy} extends this work and uses external and internal force and end-effector velocity as input to generate 6D force as output to their feed-forward force-based manipulator controller. Additionally, \cite{shuklaForceConditionedDiffusionPolicies2025, shuklaLearningForceConditionedVisuomotor} propose incorporating force into diffusion policies for manufacturing tasks.

\begin{figure}[tp]
    \centering
    \smallskip
    \vspace{1mm}
    \includegraphics[width=\linewidth]{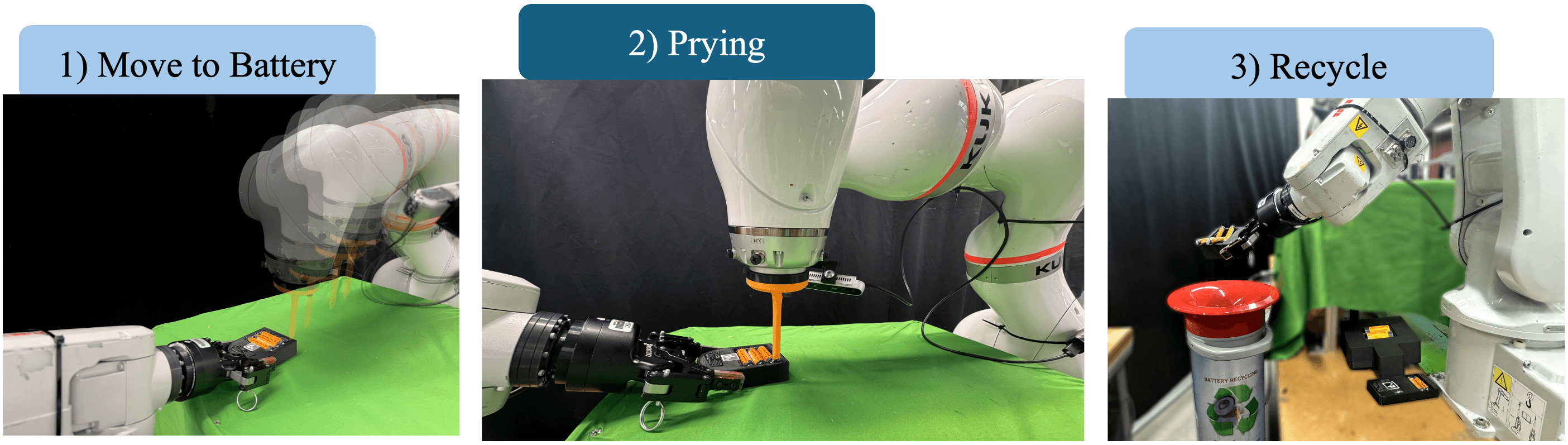}
    \caption{The battery-recycling system consists of three steps. First, the robot responsible for prying moves to the battery. Next, diffusion policy is applied to perform the prying motion. Finally, the robot holding the battery-powered product moves to the recycling bin and deposits the battery.}
    \label{fig:steps_system}

\end{figure}

\noindent \textbf{Policy Learning from Multi-sensory Input}: There has been significant research on policy learning from multi-sensory input to enhance performance compared to vision-only policies \cite{gao2022objectfolder, gao2023objectfolder}. \cite{du2022play, thankaraj2023that, mejia2024hearingtouch} use audio as extra modality with vision to map state to action and achieve higher success rate compared to benchmark methods. \cite{lee2019making} proposes fusing tactile and vision for sample-efficient policy learning for peg-in-hole problem. \cite{li2022seehearfeel} proposes combining vision, tactile, and audio as input to learn policy for robotic manipulation. Recently, \cite{liu2024maniwavlearningrobotmanipulation} proposed using image and audio as sensory observation for diffusion policy. 

In order to perform high-precision compliant object prying task, force and vision need to play a critical role in guiding policy output. However, no prior studies have addressed effectively combining lower and higher-dimensional input for robotic policy learning. This paper introduces a method to integrate force data with images to enhance action prediction using diffusion policy.

\begin{figure*}[tp]
    \centering
    \smallskip
    \includegraphics[width=\linewidth]{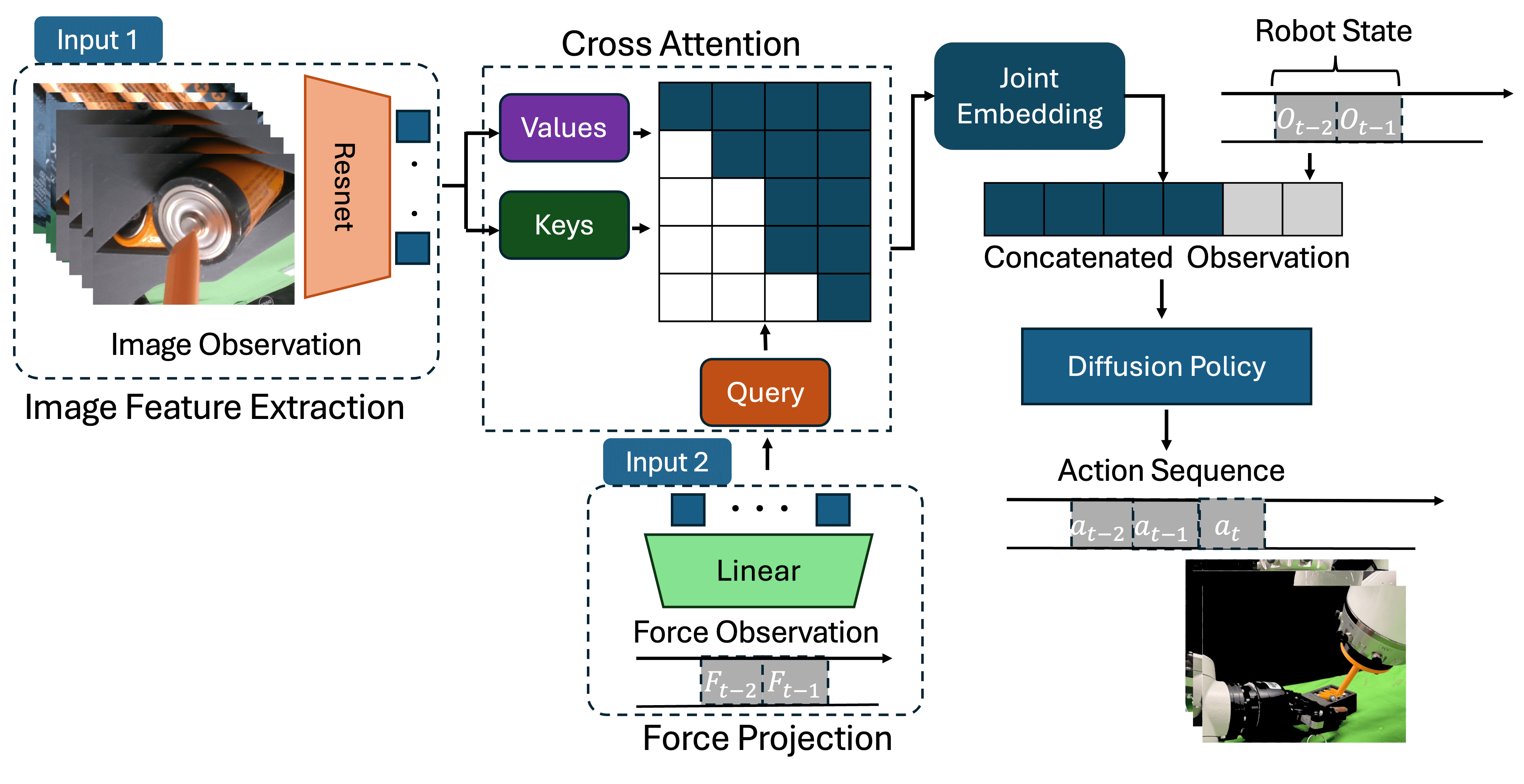}
    \caption{\textbf{Framework overview}: For image data (Input 1), ResNet \cite{he2016deep} is used to extract features and force data (Input 2) is linearly projected to match the size of the image features and is used as the query. The image is cropped to 98 × 98 (or any suitable dimensions) before being passed into ResNet. The cross-attention mechanism combines these inputs to output a joint embedding vector, which is then concatenated with the robot pose. This combined vector is incorporated into Feature-wise Linear Modulation (FiLM) conditioning \cite{perez2018film} for noise prediction within the U-Net architecture \cite{ronneberger2015u} in the diffusion framework. The output is an action sequence, $a_t$ \cite{chi2023diffusionpolicy}.}
    \label{fig:overview_arch}
    \vskip -1.5em
\end{figure*}

\section{Problem Formulation}
\label{section: diffusion_for_battery}

\noindent Consider a compliant object prying task, $\mathcal{T}$. We aim to learn a policy, $\pi$ conditioned on observation, $O_t$, at time step, $t$. Specifically, we have $O_t = \{\Gamma(I,\mathit{F}), S\}$, where $I$ denotes image, $\mathit{F}$ refers to the three-axis Cartesian force experienced in the robot's end-effector frame, and $S$ denotes 6 DoF robot end-effector state. $\Gamma(I,\mathit{F})$ refers to the joint feature embedding. The robot $\mathit{R}$ performs the prying task $\mathcal{T}$ by executing a sequence of actions $A_t$, where the action sequence is conditioned on the state observation as $P(A_t \mid O_t)$. We condition the action output on a predefined $n$ past observations, where $n$ specifies the number of previous time steps included to provide temporal context. We focus on learning the joint feature embedding $\Gamma(I,\mathit{F})$ to ensure that force information is effectively incorporated into the policy and not become underutilized as a low-dimensional input. More detailed explanation of steps involved in outputting the joint embedding is described in Section \ref{section: cross_attention}.

\noindent \textbf{Overview of Approach}: To effectively collect force data alongside vision, robot state, and action, it is essential to synchronize the timestamps at which these modalities are captured. To maximize the impact of force data on the diffusion policy, we apply data augmentation techniques to account for out-of-distribution force levels and noise that may occur during inference. The details of this force processing methodology are discussed in Section \ref{section: processing_force}. To address the dimensionality mismatch between the force and vision data, we project the force data into a higher-dimensional vector. We then employ a cross-attention architecture to capture relational features between the vision and force data. Further details of the method are provided in Section \ref{section: cross_attention}.

\noindent\textbf{Diffusion Policy Formulation}:A Denoising Diffusion Probabilistic Model (DDPM) is a generative model that learns to reverse a noise injection process through iterative denoising steps \cite{ho2020denoising}. Given a data distribution $x^K$, the model progressively denoises it through $K$ iterations, producing intermediate states, $x^k,x^{k-1}... x^0$, until the final noise-free output $x_0$ is obtained. We treat the action sequence $A_t$ as our data, and the reverse diffusion process is formulated as:

\begin{equation}
 \mathcal{\textbf{A}}_t^{k-1} = \alpha(\textbf{A}_t^k-\gamma\epsilon_\theta(O_t,A_t^k,k)+\mathcal{N}(0,\sigma^2I))
\end{equation}

Where $\epsilon_\theta$ is the noise prediction network that estimates the noise at each denoising step, $\mathcal{N}(0,\sigma^2I)$ represents the Gaussian noise injected during the forward diffusion process, $\alpha$, $\gamma$ and $\sigma$ define the noise schedule, which controls the rate of denoising and influences the learning dynamics during training.
For a more detailed explanation of the forward and reverse diffusion steps, we refer readers to \cite{ho2020denoising} and \cite{chi2024diffusionpolicy}.

\vspace{-0.6mm}

The loss function for the noise prediction network is

\begin{equation}
 \mathcal{L} = MSE(\epsilon^k, \epsilon_\theta(O_t, A^0_t+\epsilon^k,k))
\end{equation}

where $\epsilon^k$ denotes the randomly sampled noise added to the unchanged action, $A^0_t$. The diffusion policy predicts a sequence of 6-DoF delta actions, representing the relative difference between the current and the next target pose. This sequence has a length of $T_a$, the action horizon. The robot executes only a subset of actions from this sequence, corresponding to the execution horizon $T_e$.

\begin{figure}[tp]
    
    \centering
    \smallskip
    \includegraphics[width=\linewidth]{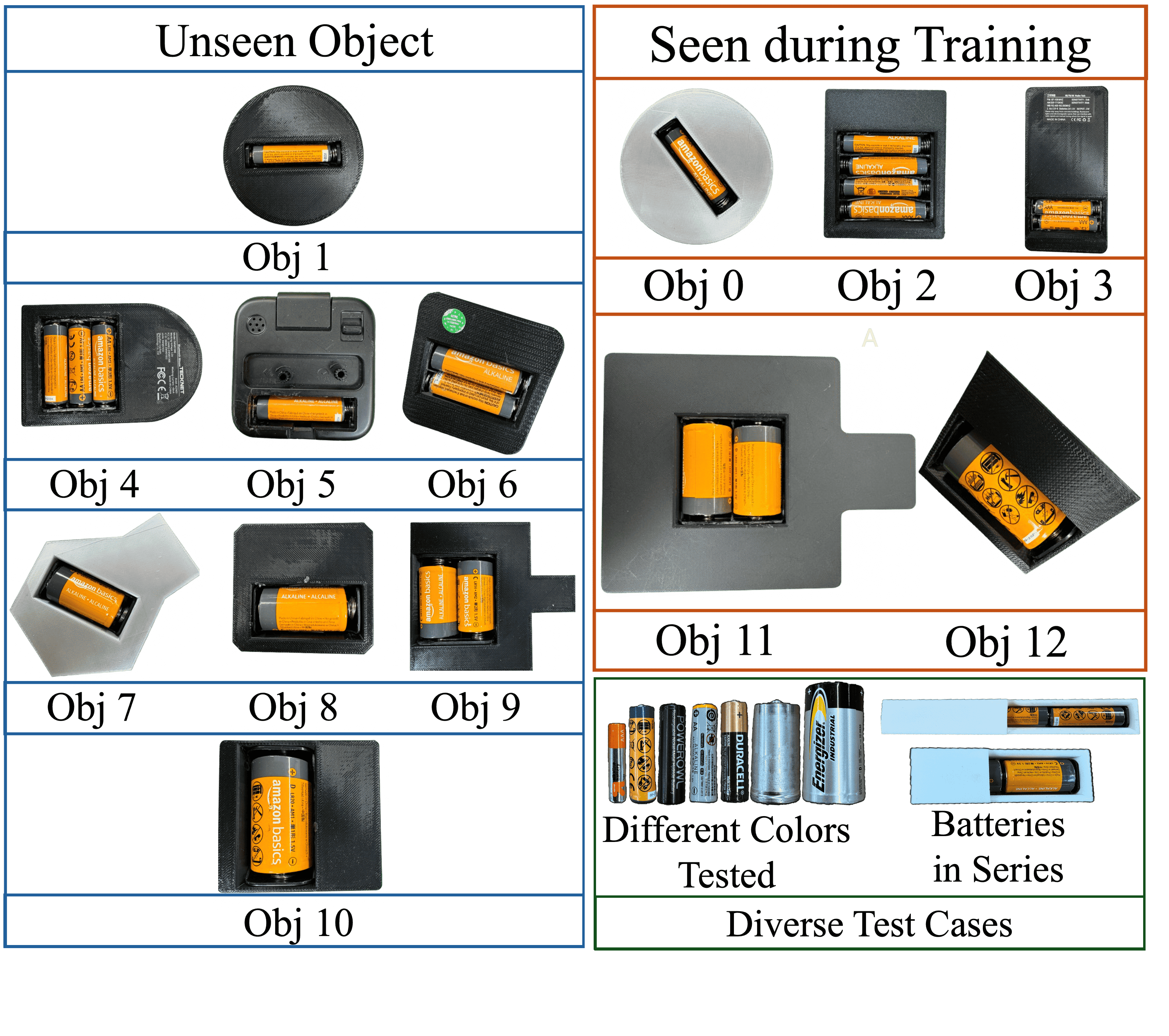}
    \caption{Products and Batteries Used in Experiments: Note that obj0 was excluded from testing to present results across three objects per battery type. Some product casings feature slanted designs with variable angles, and the depth of the casing from the top of the battery varies by approximately $\pm 4mm$. Bottom shows the products used in experiments in Section \ref{section: edge_case}.}
    \label{fig:ex_products}
\end{figure}

\section{Method}
\label{section: method}

\subsection{Processing Force for Enhanced Generalizability}
\label{section: processing_force} 

\begin{figure}[tp]
    \centering
    \smallskip
    \includegraphics[width=\linewidth]{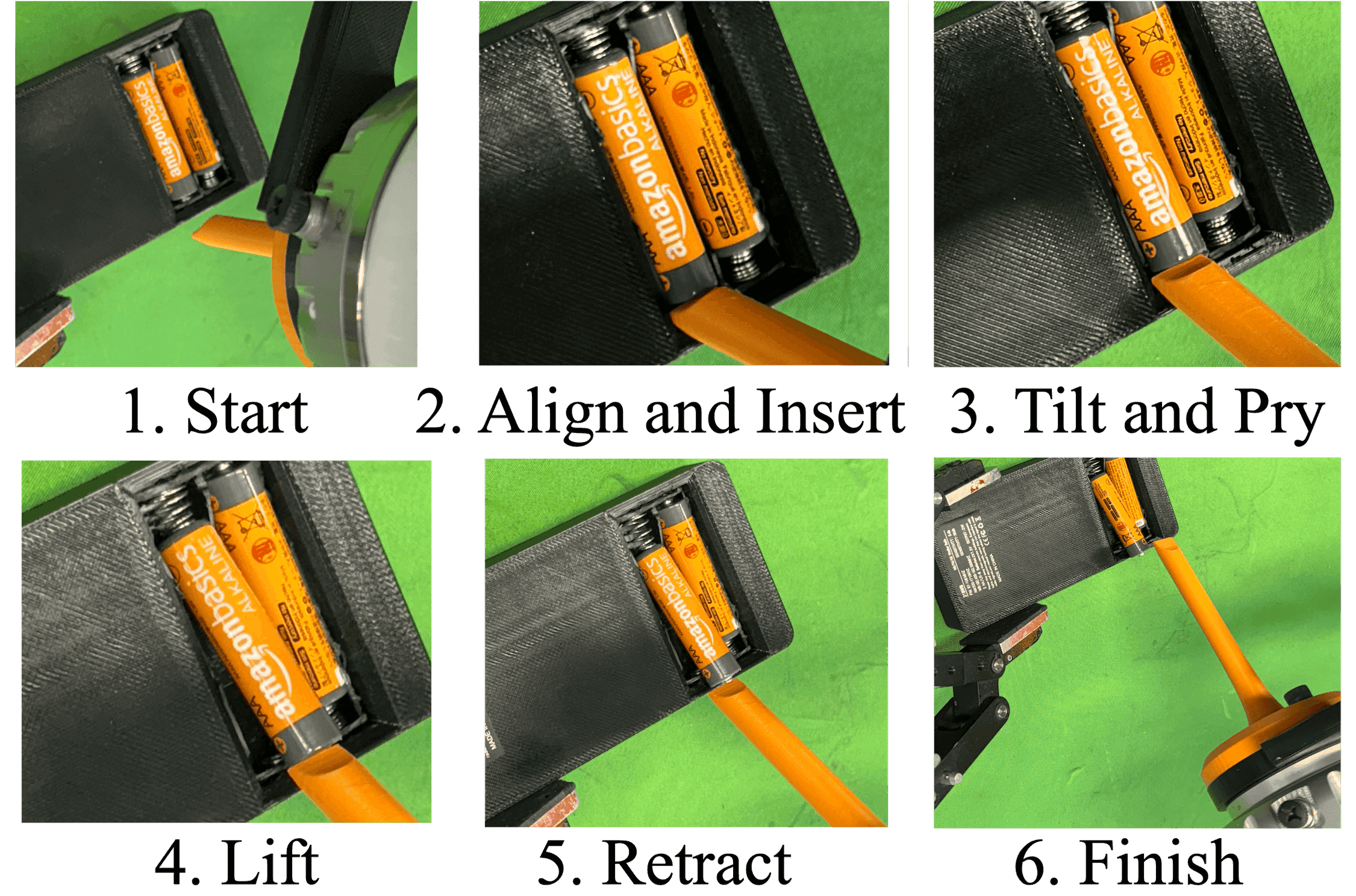}

    \caption{Steps for Prying: The robot begins from a random initial position and moves toward the battery. It then approaches the gap for insertion. Next, the robot aligns the prying tool tip with the gap at the correct angle and moves downward for insertion. Once the tool is inside the gap, the robot tilts to pry the battery. Upon applying adequate amount of force, the robot lifts the battery. Finally, it retracts, completing the task.}
    \label{fig:steps_prying}
\end{figure}

\noindent \textbf{Synchronizing Observation Modalities}: Studies have explored improving policy performance by synchronizing observation latencies across different modalities \cite{chi2024universal}. For tasks that do not involve dynamic movements and use relatively low sampling rates (under 10Hz), we find that precise synchronization under 0.1 seconds is unnecessary. However, aligning force data with other modalities like image, robot state, and action remains crucial for accurate inference. In our approach, we simultaneously sample force, robot state, and image data at each action execution step, achieving sub-half-second sampling precision. For tasks with significantly less frequent sampling rate for force, we recommend readers to the force interpolation techniques described in \cite{wu2024tacdiffusionforcedomaindiffusionpolicy}.

\noindent \textbf{Force Data Augmentation}: A major challenge in compliant object prying is the high variability in force needed. To generalize better to out-of-distribution objects, it is essential to inform the model that different maximum force levels can still achieve prying. To address this issue, we introduce random scaling and Gaussian noise to the force data during training, allowing adaptation to varying force levels across objects with different rigidity. We scale the force data by a factor uniformly sampled from $[0.9, 1.2]$ and add Gaussian noise $\mathcal{N}(0,0.005)$ to simulate noise during inference. We adopt a higher upper bound for force scaling because lower force applied often result in task failure, whereas applying a reasonably higher force tends to improve the likelihood of successful prying.

\subsection{Using Cross Attention to Learn Relational Features between Image and Force}
\label{section: cross_attention} 

\noindent During our exploration, we found that simply concatenating Cartesian force components with image features was insufficient for effectively conditioning action output. We hypothesize that the higher-dimensional image features overshadow the lower-dimensional force data. To address this, we initially scaled the force input using MLP or linear layers to match the dimensionality of the image features. However, as shown in Section \ref{section: results}, this approach did not lead to significant performance improvements.



To learn $\Gamma(I,F)$, we adopt a cross-attention mechanism to capture relational features between RGB images and force data. Cross-attention has been proven effective in a variety of tasks, including language translation \cite{ashish_attention_2017}, where it excels at learning contextual mappings between input sequences. Cross-attention can capture complex relationships and dependencies by attending to relevant parts of one input while processing another. In our application, we leverage cross-attention to learn a joint embedding between heterogeneous modalities that guide the policy output in diffusion policy. This mechanism enables the model to focus on the features of one modality while dynamically integrating complementary information from another.

We use ResNet-18 to encode the image data into a lower-dimensional feature representation with $N$ spatial features and $d$ feature dimensions, resulting in $I_{encoded} \in R^{N\times d}$. To enable the neural network to process force information effectively while preserving all relevant data, we separate the force into two components: magnitude and direction. The final input to the network consists of four parameters: the normalized magnitude and three directional components, $({|F|},\hat{\mathbf{F}})$.
Subsequently, we apply a linear projection layer to expand the force input to match the $I_{encoded}$. Let the force feature be $F_{projected} \in R^{4 \times d}$. Then in cross-attention mechanism, we have projection matrices $W^K, W^V, W^Q$, and leads to key, value and query input

\begin{equation}
\begin{aligned}
    K &= I_{\text{encoded}} \cdot W^K \\
    V &= I_{\text{encoded}} \cdot W^V\\
    Q &= F_{\text{projected}} \cdot W^Q \\
\end{aligned}
\end{equation}

Then finally the matrix output is computed \cite{ashish_attention_2017}.

\begin{equation}
\mathit{\textbf{A} = softmax(\frac{\textbf{Q}\cdot\textbf{K}^T}{\sqrt{d}})\textbf{V}} 
\end{equation}

where $d$ refers to the dimension of the key/query vectors for numerical stability \cite{ashish_attention_2017}. In our architecture, we use hidden dimension of 512 and 4 attention heads. Finally, a two-layer MLP refines the output. We experimentally find that using force as the query source outperforms an architecture using it as key and value. Full architecture of the learning framework is shown in Fig. \ref{fig:overview_arch}.

\section{Experimental Setup}
\label{section: training} 

\subsection{Testbed}
\noindent We have a bi-manual setup in which one robot is responsible for extracting the battery from the casing, and the other transports the extracted batteries from the workstation to the recycling bin. We use a KUKA IIWA 14 for battery extraction and an ABB IRB120 for support arm. The diffusion policy only includes the extraction robot. Figures of the setup can be found in Fig. \ref{fig:steps_system}.

We test our method on various types of battery-powered products as shown in Fig. \ref{fig:ex_products}. We have four types of batteries: AAA, AA, C, and D, which are commonly used for household products. These variations in size of the batteries introduce different challenges in prying out the batteries from the casing such as precision during insertion, force required while pushing against the spring, and traction while lifting up the battery. During our experiment, we vary position and orientation of the object and initial robot positions to test the model's robustness to different initial object configuration.

We treat each battery removal as a separate episode, allowing each prying task to be handled individually and yet still maintaining the whole model as one. Therefore, we use traditional methods like Aruco markers or object detection to position the end-effector near one end of the battery, then start diffusion-based inference for prying as shown in Fig. \ref{fig:steps_system}. For our experiments, we start within an imaginary $1.0 cm \times 1.0 cm \times 2.0 cm$ area (allowing more variability in $z$), accounting for typical error margins of these approaches.

\subsection{Data Collection}

\noindent For data collection, we have an expert human demonstrator hand-guide the robot to perform the task at the rate of $3Hz$. However, inherently, hand-guiding requires humans to move the robot directly. This requirement makes it incompatible with the diffusion policy with images since humans are captured during RGB data collection. To avoid this issue, once human demonstration is done, we replay the recorded trajectory to extract force, robot state, action, and image data, enabling validation of the quality of the human-demonstrated trajectory. Although rare incidents of replay failure can occur due to perturbations in the initial conditions caused by the force exerted during the human demonstration, we mitigated this issue by exercising caution during demonstrations. This significantly improved the success rate, with failures ranging from zero to two observed out of thirty replays. For image data, we use a RealSense D415 camera mounted on the robot's wrist, and for force data, we rely on the built-in force/torque sensor of the KUKA IIWA14 robot. Notably, any multi-axis force/torque sensor can be used for this setup.
\vskip -1em
\subsection{Training}
\noindent During training, we use AAA and D battery, with three types of casing for AAA and two for type-D as shown in Fig. \ref{fig:ex_products}. There are 419 episodes of demonstration, where each consists of single battery removal. For state and action definition, we use 6D representation for continuous rotation representation \cite{Zhou_2019_CVPR}. We find that using delta action representation yields better results compared to absolute representation. Absolute actions tend to over-fit to specific positions and orientations.

Our Resnet-18 vision backbone is trained end-to-end. Our experiments show that training the model end-to-end on our dataset performs better than fine-tuning a pre-trained model. This is likely because end-to-end training allows the network to learn task-specific feature representations more effectively.

We use the global average pooling combined with spatial softmax pooling to preserve spatial information and replace Batch Normalization (BatchNorm) with Group Normalization (GroupNorm) for training stability. Additionally, we incorporate the Exponential Moving Average (EMA) during training. Key hyper-parameters include a learning rate of $2 \times 10^{-4}$, a weight decay of $1 \times 10^{-3}$, and a diffusion step count of 100. We
use observation count of, $n = 2$, and action horizon of 16,
of which only 6 actions are executed to allow continuous
re-plan suited for high-precision tasks. For image, we use color jitter and random crop as done in \cite{chi2024universal, liu2024maniwavlearningrobotmanipulation}. For the parameter for color jitter, we use brightness 0.4, contrast 0.4, saturation 0.2, and hue 0.1. We then train the model for 2000 epochs with batch size of 72. We train our models on Nvidia RTX 3080 GPU for around 14 hours.

\section{Results}
\label{section: results}
\subsection{Success Rate Test}
\label{section: success_rate_test}
\noindent In order to demonstrate that each component in our architecture produces boost in the performance of the task, we benchmark our method against three different methods:

\begin{enumerate}
    \item \textbf{DP-B (Baseline Diffusion Policy)}  : Vision-only diffusion policy \cite{chi2024diffusionpolicy}
    \item \textbf{DP-LF (Diffusion Policy with Low-Dimensional Force)}: Diffusion Policy with image conditioning and low-dimensional force, vector of size four (see Section \ref{section: processing_force}) concatenated to image features 
    \item \textbf{DP-PF (Diffusion Policy with Projected Force)}: Diffusion Policy with image and force feature up-scaled using linear projection to match image dimension 
    \item \textbf{DP-CA (Diffusion Policy with Cross-Attention between Image and Force)} (Ours) Diffusion Policy with learned joint embedding between image and force using cross attention (see Section \ref{section: cross_attention})
\end{enumerate}

We test our method on 12 different objects (see Fig. \ref{fig:ex_products}). Each object has a distinct shape with varying angles, colors, tolerances, and depths. We focus on the depth range found in most single-layered products. For objects with deeper casings, a sharper metal tool could be designed with appropriate safety measures in place to prevent accidental battery puncture. We conduct 10 inference iterations per object, total of 120 experiments for each model. In \cite{chi2024diffusionpolicy}, it has already been seen that for multi-modal tasks, diffusion policy outperforms other behavior cloning approaches like LSTM-GMM \cite{mandlekar2020iris}, IBC \cite{florence2022implicit}, and BET \cite{shafiullah2022behavior}. Therefore, we did not consider these approaches in our benchmark set.  

We define success as prying out and lifting the battery to a point where, if the secondary arm tilts the product, the batteries will fall entirely into the recycling bin. When this prying task fully loosens the battery, it is deemed successful. Steps for prying out the battery are depicted in Fig. \ref{fig:steps_prying}.

Table \ref{tab:success_rate_combined} shows the comparison in success rate among benchmark methods and our method. As shown in the result, we see that our method outperforms the vision-only baseline \cite{chi2023diffusionpolicy} by $57\%$. Additionally, we outperform other variant methods that use force by $48\%$ and $39\%$. Particularly, this significant difference stems from the unseen objects, which shows that adding joint embedding between vision and image can enhance generalizability among unseen objects more than any other methods of using force. It is also worth noting that using force in some naive way as done in method \textbf{DP-LF} and \textbf{DP-PF} improves the success rate compared to vision-only baseline by $9\%$ to $18\%$, but it still struggles to generalize to objects not seen during training.

\noindent \textit{Failure Mode Analysis}: During inference, our method primarily showed failure modes while prying out AAA batteries, as shown in Tab. \ref{tab:success_rate_combined}. These failures resulted from the tight tolerances of AAA batteries, which reduced the contact area between the prying tool and the battery. Although the robot generally succeeded in inserting the tooltip correctly, most failures were caused by a slight misalignment between the tool and the gap. For the single failure with an AA battery in object 5, we categorized it as a failure because the robot failed to fully lift the battery before retracting, leaving the battery partially pried out. 

In other baseline methods, failure modes include failure to insert the tool, premature prying before full insertion, insufficient force during prying, and insufficient contact while lifting the battery. We direct readers to the project website, where videos of the failure cases are available.

\begin{table*}[h!]
    \centering
    \vspace{0.5cm}
    \renewcommand{\arraystretch}{1.3} 
    \large 
    \resizebox{\textwidth}{!}{
    \begin{tabular}{@{}l!{\vrule width 0.5pt}cccc!{\vrule width 0.5pt}cccc!{\vrule width 0.5pt}cccc!{\vrule width 0.5pt}cccc!{\vrule width 0.5pt}c@{}}
        \toprule
        & \multicolumn{4}{c!{\vrule width 0.5pt}}{ AAA} & \multicolumn{4}{c!{\vrule width 0.5pt}}{ AA} & \multicolumn{4}{c!{\vrule width 0.5pt}}{ C} & \multicolumn{4}{c!{\vrule width 0.5pt}}{ D} &  Avg. Success Rate \\
        \cmidrule(lr){2-5} \cmidrule(lr){6-9} \cmidrule(lr){10-13} \cmidrule(lr){14-17} \cmidrule(l){18-18}
         & Obj1 & Obj2* & Obj3* & {Avg.} & Obj4 & Obj5 & Obj6 & {Avg.} & Obj7 & Obj8 & Obj9 & {Avg.} & Obj10 & Obj11* & Obj12* & {Avg.} & Overall Avg. \\ 
        \midrule
        DP-B & 0.0 & 0.5 & 0.7 & 0.4 & 0.2 & 0.3 & 0.1 & 0.2 & 0.4 & 0.4 & 0.3 & 0.37 & 0.4 & 0.8 & 0.6 & 0.6 & 0.39 \\ 
        DP-LF & 0.2 & 0.4 & 0.7 & 0.43 & 0.5 & 0.4 & 0.3 & 0.4 & 0.1 & 0.4 & 0.8 & 0.43 & 0.6 & 0.5 & 0.9 & 0.67 & 0.48 \\ 
        DP-PF & 0.4 & 0.7 & 0.8 & 0.63 & 0.5 & 0.7 & 0.5 & 0.57 & 0.2 & 0.3 & 0.7 & 0.4 & 0.4 & 0.7 & 0.9 & 0.67 & 0.57 \\ 
        \textbf{DP-CA\newline 
        (Ours)} & \textbf{0.9} & \textbf{0.9} & \textbf{0.9} & \textbf{0.9} & \textbf{1.0} & \textbf{0.9} & \textbf{1.0} & \textbf{0.97} & \textbf{1.0} & \textbf{1.0} & \textbf{1.0} & \textbf{1.0} & \textbf{1.0} & \textbf{1.0} & \textbf{1.0} & \textbf{1.0} & \textbf{0.96} \\
        \bottomrule
    \end{tabular}
    }
    \caption{Success Rate Comparison Among In-Distribution and Out-of-Distribution Objects and Battery Types: This comparison includes individual object performance, with * denoting in-distribution objects. Section \ref{section: success_rate_test} details the benchmarking methods. Our method outperforms the vision-only model by 57\% and surpasses other force-incorporated benchmarks by $48\%$ and $39\%$. Notably, our method performs marginally better on AA and C-type batteries, which were unseen during training, demonstrating its generalizability among unseen objects.}
    \label{tab:success_rate_combined}
    \vskip -1.5em

\end{table*}

{\subsection{Performance under Edge Cases}
\label{section: edge_case}

This section evaluates our model under edge cases not covered in the previous section and unseen during training. Specifically, we test two scenarios: prying batteries connected in series (two or three batteries) and batteries with varying colors. We perform 20 experiments for each category using the success criteria from Section \ref{section: success_rate_test}. For batteries in series, we use two or three type-C and AA batteries in white casings. For varying colors, we test six colors—brown, black, dark orange, bright orange, metallic (reflective), and grey—across four battery types as shown in Figure \ref{fig:ex_products}}

The model achieves a $90\%$ success rate for batteries with varying colors and $95\%$ for batteries in series, demonstrating a success rate comparable to Table \ref{tab:success_rate_combined}. These results indicate that our model can generalize to unseen diverse scenarios.

\subsection{Comparison in Time Taken between Human Demonstration and Robot Inference}
\noindent We compare the time a human takes to perform the prying task by hand-guiding the robot against the robot's inference time. We record 30 iterations per battery type, measuring from the start until the robot fully removes the battery from the casing.

\begin{figure}
    \centering
    \vspace{1mm}
    \noindent\includegraphics[width=\linewidth]{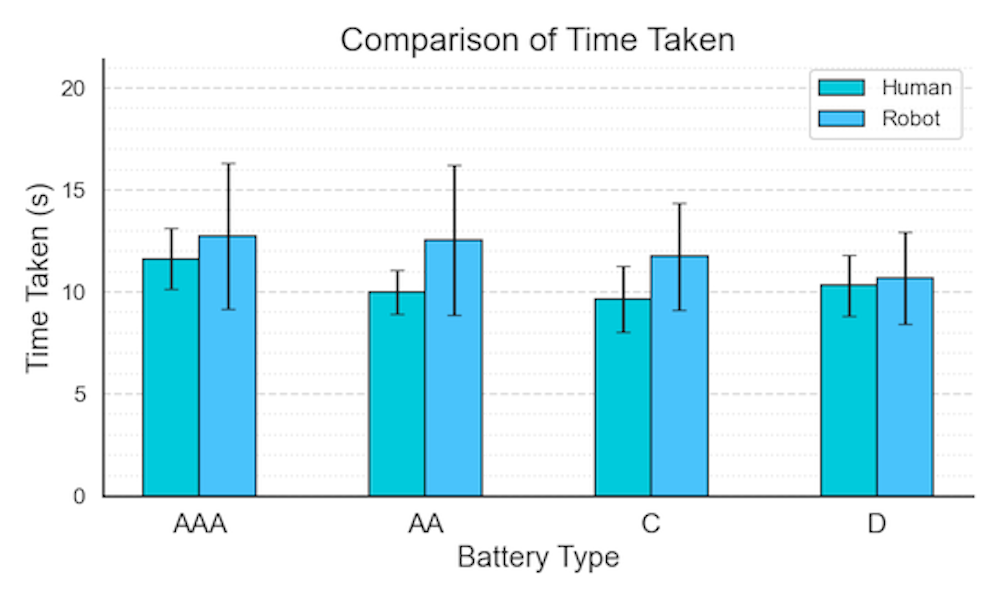}
    \caption{A comparison of the average time taken for the human demonstration via kinesthetic teaching versus robot inference.}
    \label{fig:time_comparison}
\end{figure}

\begin{figure}
    \centering
    \vspace{1mm}

    \noindent\includegraphics[width=    \linewidth]{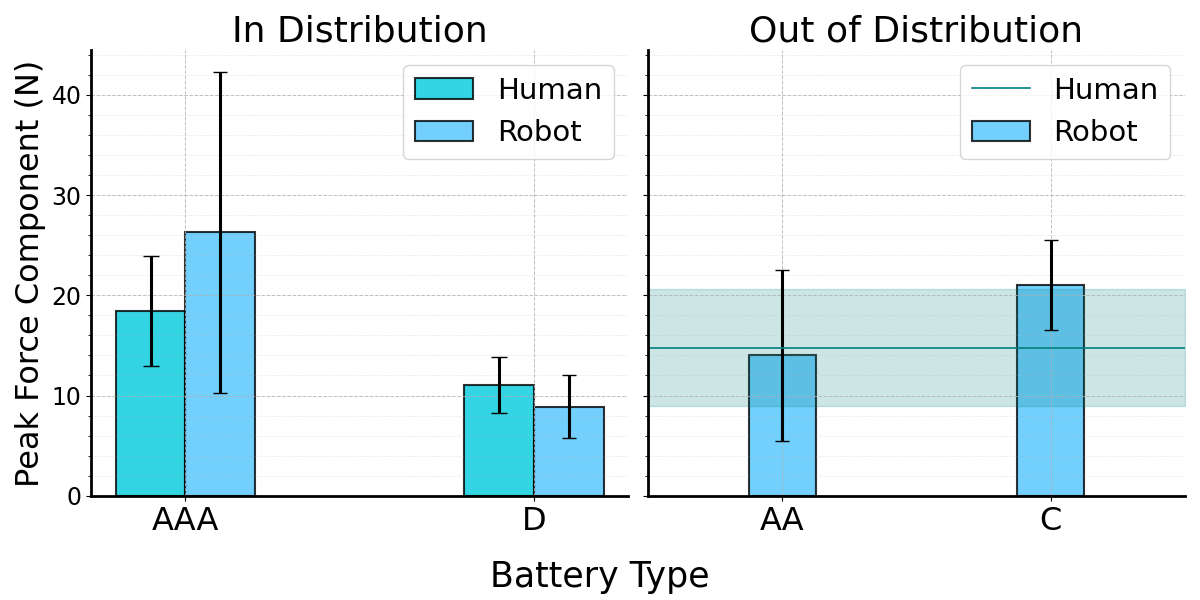}
    \caption{Comparison between peak component force exerted on batteries between the human demonstration and robot inference. Since AA and C are out of distribution, we show the average of force during human demonstration exerted on type AAA and D.}
    \label{fig:max_force}
    \vskip -0.5em
\end{figure}

Fig. \ref{fig:time_comparison} shows the average time comparison between human guidance and robotic inference. This metric shows that our learning from demonstration framework can perform tasks comparatively to the rate at which the demonstration was provided, though the robot requires slightly more time. This additional time is due to occasional idle actions in the beginning of inference and the approximate 1-second duration of each inference step. We observe that performing battery prying with bare hands would take significantly less time. However, this gap in execution time can be addressed by increasing the speed of the demonstration, raising the manipulator's velocity, and tuning impedance values. We opted not to implement these adjustments to prioritize safety and avoid potential risks related to battery damage.

\vspace{-0.5em}

\begin{figure}
\begin{center}
    \includegraphics[width=\linewidth] {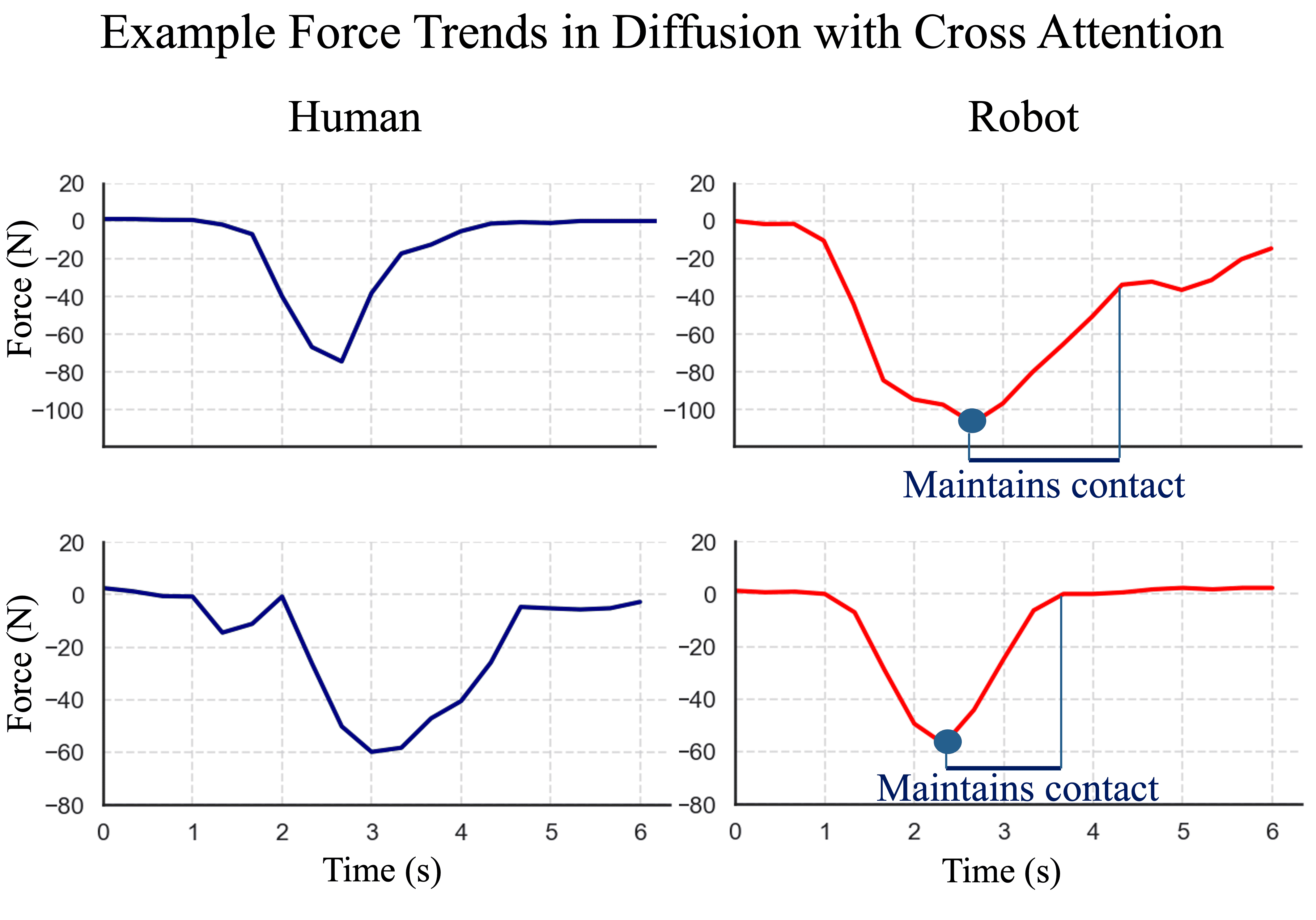}
    \caption{Force Trend During Battery Prying Task: Force trend during robot inference closely aligns with the trend in the human demonstration. Success in the prying task relies on maintaining contact throughout the prying and lifting phases: see Fig. \ref{fig:steps_prying}.}
    \label{fig: force_line_chart}
\end{center}
\vskip -0.7em

\end{figure}

\subsection{Comparison of Force Applied to the Battery during Human Demonstration and Robot Inference}
\noindent To demonstrate that the robot’s actions and resultant force align with the demonstration data, we compare the maximum force applied to the battery by both the human and the robot. Our experimental observations show that the maximum force in the z-component is crucial for task success. Fig. \ref{fig:max_force} shows that the maximum force applied by the robot is comparable to that of the human during the demonstration. Maintaining this force is important because insufficient force often leads to loss of contact with the battery, causing failure, while excessive force may break the tooltip or damage the battery.

\begin{figure}
\begin{center}
    \includegraphics[width=\linewidth] {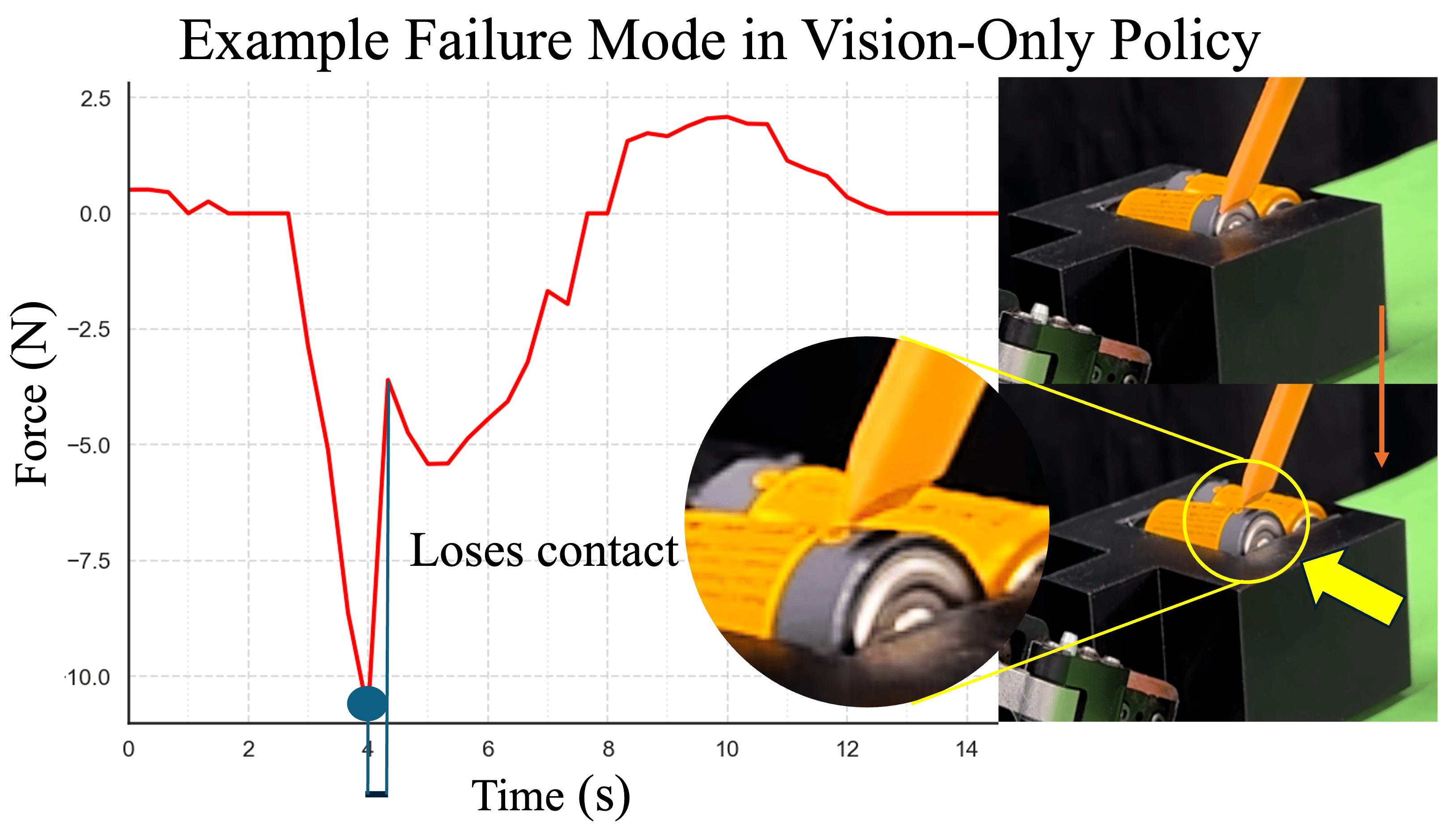}
    \caption{Example Force Trend During Failure [Vision-only]: The force curve shows a distinct difference in the force trend during inference with our method shown in Fig. \ref{fig: force_line_chart}. This curve underscores the importance of force feedback for successful execution in prying.}
    \label{fig: failure_force}
\end{center}
\vskip -0.7em
\end{figure}

We also find that force modality helps the robot detect state changes, guiding it to different action modes like insertion and prying, as illustrated in Fig. \ref{fig:steps_prying}. Our method establishes relevant features between vision and force, allowing force data to signal when to start prying if image features alone are insufficient. This additional information significantly improves the success rate, as failures in benchmark methods often stem from premature prying before full insertion. Fig. \ref{fig: force_line_chart} demonstrates that the maximum force and its trend during inference closely align with the human demonstration. Additionally, Fig. \ref{fig: failure_force} presents the force trend in a failure case during inference with the vision-only diffusion policy, illustrating the impact of missing force data on task success.

While we include force as an observation, additional measures are needed to ensure that the actions generated do not produce forces exceeding those applied during demonstrations. While one could design a controller to apply a maximum external force limit, this may reduce success rates as it may drive the planned trajectory to deviate from the diffusion-generated trajectory. We hope to address this in our future work by focusing on enforcing force thresholds without compromising the success of the task. 

\subsection{Key Findings}
\noindent \textit{Force enhances robustness in contact-rich tasks.} This result aligns with our expectations, as force data at each time step enables the robot to maintain the necessary force to ensure traction between the tool and battery tips.

\noindent \textit{Relational Feature between Image and Force improves generalization in diffusion policy when handling out-of-distribution object sizes and configurations (e/g., battery types).} During success rate testing (see Table \ref{tab:success_rate_combined}), we observe that the benchmark diffusion policies struggle significantly with unseen battery types and objects. This can be attributed to the fact that force trends and relational features remain more consistent than vision-only across different test products, allowing force to serve as an in-distribution modality that informs the robot's state alongside RGB information.

\noindent \textit{Cross-attention between image and force provides robustness in mode transitions.} This added relational insight enables the model to infer when to initiate prying, thus avoiding premature or incorrect actions such as prying on the product casing, often seen in benchmark methods. Our method provides critical information linking image features with force values, resulting in more informed mode changes during multi-step planning.

\section{Conclusions}
\label{section:conclusions}
\noindent We present an approach to effectively incorporate force as a modality alongside images in diffusion policy, specifically for prying compliant objects. A simple concatenation of low-dimensional force data with high-dimensional image features often weakens the conditioning in the noise prediction network, and straightforward force projection adds minimal performance improvement. To address this, we propose using cross-attention between image and force modalities to learn relational features, enhancing the synergy between vision and force data. Trained on two battery types, AAA and D, we achieve a 96\% success rate on unseen battery types, 57\% improvement over the vision-only baseline. Our model also shows generalization to diverse test scenarios such as unseen color and battery configurations. Additionally, metrics such as time taken, force trends, and maximum force applied match human expert demonstrations.

While we use battery prying as an example of a contact-rich task requiring force feedback, the proposed framework has the potential to improve success rates in other contact-rich tasks, similar to how \cite{chi2024diffusionpolicy} extended vision-only policies. In future work, our objective is to explore additional task scenarios to further demonstrate the applicability of our approach and conduct a comprehensive study that compares its performance with the baseline methods.

\bibliographystyle{IEEEtran}

\begin{thebibliography}{10}
\providecommand{\url}[1]{#1}
\csname url@samestyle\endcsname
\providecommand{\newblock}{\relax}
\providecommand{\bibinfo}[2]{#2}
\providecommand{\BIBentrySTDinterwordspacing}{\spaceskip=0pt\relax}
\providecommand{\BIBentryALTinterwordstretchfactor}{4}
\providecommand{\BIBentryALTinterwordspacing}{\spaceskip=\fontdimen2\font plus
\BIBentryALTinterwordstretchfactor\fontdimen3\font minus \fontdimen4\font\relax}
\providecommand{\BIBforeignlanguage}[2]{{%
\expandafter\ifx\csname l@#1\endcsname\relax
\typeout{** WARNING: IEEEtran.bst: No hyphenation pattern has been}%
\typeout{** loaded for the language `#1'. Using the pattern for}%
\typeout{** the default language instead.}%
\else
\language=\csname l@#1\endcsname
\fi
#2}}
\providecommand{\BIBdecl}{\relax}
\BIBdecl

\bibitem{fang2019survey}
B.~Fang, S.~Jia, D.~Guo, M.~Xu, S.~Wen, and F.~Sun, ``Survey of imitation learning for robotic manipulation,'' \emph{International Journal of Intelligent Robotics and Applications}, vol.~3, pp. 362--369, 2019.

\bibitem{argall2009survey}
B.~D. Argall, S.~Chernova, M.~Veloso, and B.~Browning, ``A survey of robot learning from demonstration,'' \emph{Robotics and autonomous systems}, vol.~57, no.~5, pp. 469--483, 2009.

\bibitem{chi2023diffusionpolicy}
C.~Chi, S.~Feng, Y.~Du, Z.~Xu, E.~Cousineau, B.~Burchfiel, and S.~Song, ``Diffusion policy: Visuomotor policy learning via action diffusion,'' in \emph{Proceedings of Robotics: Science and Systems (RSS)}, 2023.

\bibitem{chi2024diffusionpolicy}
C.~Chi, Z.~Xu, S.~Feng, E.~Cousineau, Y.~Du, B.~Burchfiel, R.~Tedrake, and S.~Song, ``Diffusion policy: Visuomotor policy learning via action diffusion,'' \emph{The International Journal of Robotics Research}, 2024.

\bibitem{chi2024universal}
C.~Chi, Z.~Xu, C.~Pan, E.~Cousineau, B.~Burchfiel, S.~Feng, R.~Tedrake, and S.~Song, ``Universal manipulation interface: In-the-wild robot teaching without in-the-wild robots,'' in \emph{Proceedings of Robotics: Science and Systems (RSS)}, 2024.

\bibitem{liu2024maniwavlearningrobotmanipulation}
\BIBentryALTinterwordspacing
Z.~Liu, C.~Chi, E.~Cousineau, N.~Kuppuswamy, B.~Burchfiel, and S.~Song, ``Maniwav: Learning robot manipulation from in-the-wild audio-visual data,'' 2024. [Online]. Available: \url{https://arxiv.org/abs/2406.19464}
\BIBentrySTDinterwordspacing

\bibitem{10378967}
P.~So, A.~Sarabakha, F.~Wu, U.~Culha, F.~J. Abu-Dakka, and S.~Haddadin, ``Digital robot judge: Building a task-centric performance database of real-world manipulation with electronic task boards,'' \emph{IEEE Robotics \& Automation Magazine}, vol.~31, no.~4, pp. 32--44, 2024.

\bibitem{bain1995framework}
M.~Bain and C.~Sammut, ``A framework for behavioural cloning.'' in \emph{Machine Intelligence 15}, 1995, pp. 103--129.

\bibitem{zhang2018deep}
T.~Zhang, Z.~McCarthy, O.~Jow, D.~Lee, X.~Chen, K.~Goldberg, and P.~Abbeel, ``Deep imitation learning for complex manipulation tasks from virtual reality teleoperation,'' in \emph{2018 IEEE international conference on robotics and automation (ICRA)}.\hskip 1em plus 0.5em minus 0.4em\relax IEEE, 2018, pp. 5628--5635.

\bibitem{mandlekar2020learning}
A.~Mandlekar, D.~Xu, R.~Mart{\'\i}n-Mart{\'\i}n, S.~Savarese, and L.~Fei-Fei, ``Learning to generalize across long-horizon tasks from human demonstrations,'' \emph{arXiv preprint arXiv:2003.06085}, 2020.

\bibitem{zeng2021transporter}
A.~Zeng, P.~Florence, J.~Tompson, S.~Welker, J.~Chien, M.~Attarian, T.~Armstrong, I.~Krasin, D.~Duong, V.~Sindhwani \emph{et~al.}, ``Transporter networks: Rearranging the visual world for robotic manipulation,'' in \emph{Conference on Robot Learning}.\hskip 1em plus 0.5em minus 0.4em\relax PMLR, 2021, pp. 726--747.

\bibitem{seita2020deep}
D.~Seita, A.~Ganapathi, R.~Hoque, M.~Hwang, E.~Cen, A.~K. Tanwani, A.~Balakrishna, B.~Thananjeyan, J.~Ichnowski, N.~Jamali \emph{et~al.}, ``Deep imitation learning of sequential fabric smoothing from an algorithmic supervisor,'' in \emph{2020 IEEE/RSJ International Conference on Intelligent Robots and Systems (IROS)}.\hskip 1em plus 0.5em minus 0.4em\relax IEEE, 2020, pp. 9651--9658.

\bibitem{rahmatizadeh2018vision}
R.~Rahmatizadeh, P.~Abolghasemi, L.~B{\"o}l{\"o}ni, and S.~Levine, ``Vision-based multi-task manipulation for inexpensive robots using end-to-end learning from demonstration,'' in \emph{2018 IEEE international conference on robotics and automation (ICRA)}.\hskip 1em plus 0.5em minus 0.4em\relax IEEE, 2018, pp. 3758--3765.

\bibitem{florence2019self}
P.~Florence, L.~Manuelli, and R.~Tedrake, ``Self-supervised correspondence in visuomotor policy learning,'' \emph{IEEE Robotics and Automation Letters}, vol.~5, no.~2, pp. 492--499, 2019.

\bibitem{schulman2016learning}
J.~Schulman, J.~Ho, C.~Lee, and P.~Abbeel, ``Learning from demonstrations through the use of non-rigid registration,'' in \emph{Robotics Research: The 16th International Symposium ISRR}.\hskip 1em plus 0.5em minus 0.4em\relax Springer, 2016, pp. 339--354.

\bibitem{florence2022implicit}
P.~Florence, C.~Lynch, A.~Zeng, O.~A. Ramirez, A.~Wahid, L.~Downs, A.~Wong, J.~Lee, I.~Mordatch, and J.~Tompson, ``Implicit behavioral cloning,'' in \emph{Conference on Robot Learning}.\hskip 1em plus 0.5em minus 0.4em\relax PMLR, 2022, pp. 158--168.

\bibitem{jarrett2020strictly}
D.~Jarrett, I.~Bica, and M.~van~der Schaar, ``Strictly batch imitation learning by energy-based distribution matching,'' \emph{Advances in Neural Information Processing Systems}, vol.~33, pp. 7354--7365, 2020.

\bibitem{du2019implicit}
Y.~Du and I.~Mordatch, ``Implicit generation and modeling with energy based models,'' \emph{Advances in Neural Information Processing Systems}, vol.~32, 2019.

\bibitem{shafiullah2022behavior}
N.~M. Shafiullah, Z.~Cui, A.~A. Altanzaya, and L.~Pinto, ``Behavior transformers: Cloning $ k $ modes with one stone,'' \emph{Advances in neural information processing systems}, vol.~35, pp. 22\,955--22\,968, 2022.

\bibitem{mandlekar2020iris}
A.~Mandlekar, F.~Ramos, B.~Boots, S.~Savarese, L.~Fei-Fei, A.~Garg, and D.~Fox, ``Iris: Implicit reinforcement without interaction at scale for learning control from offline robot manipulation data,'' in \emph{2020 IEEE International Conference on Robotics and Automation (ICRA)}.\hskip 1em plus 0.5em minus 0.4em\relax IEEE, 2020, pp. 4414--4420.

\bibitem{sohl2015deep}
J.~Sohl-Dickstein, E.~Weiss, N.~Maheswaranathan, and S.~Ganguli, ``Deep unsupervised learning using nonequilibrium thermodynamics,'' in \emph{International conference on machine learning}.\hskip 1em plus 0.5em minus 0.4em\relax PMLR, 2015, pp. 2256--2265.

\bibitem{ho2020denoising}
J.~Ho, A.~Jain, and P.~Abbeel, ``Denoising diffusion probabilistic models,'' \emph{Advances in neural information processing systems}, vol.~33, pp. 6840--6851, 2020.

\bibitem{wang2023diffusion}
Z.~Wang, J.~J. Hunt, and M.~Zhou, ``Diffusion policies as an expressive policy class for offline reinforcement learning,'' in \emph{The Eleventh International Conference on Learning Representations}, 2023.

\bibitem{reuss2023goal}
M.~Reuss, M.~Li, X.~Jia, and R.~Lioutikov, ``Goal-conditioned imitation learning using score-based diffusion policies,'' \emph{arXiv preprint arXiv:2304.02532}, 2023.

\bibitem{pearce2023imitating}
T.~Pearce, T.~Rashid, A.~Kanervisto, D.~Bignell, M.~Sun, R.~Georgescu, S.~V. Macua, S.~Z. Tan, I.~Momennejad, K.~Hofmann \emph{et~al.}, ``Imitating human behaviour with diffusion models,'' \emph{arXiv preprint arXiv:2301.10677}, 2023.

\bibitem{urain2023se}
J.~Urain, N.~Funk, J.~Peters, and G.~Chalvatzaki, ``Se (3)-diffusionfields: Learning smooth cost functions for joint grasp and motion optimization through diffusion,'' in \emph{2023 IEEE International Conference on Robotics and Automation (ICRA)}.\hskip 1em plus 0.5em minus 0.4em\relax IEEE, 2023, pp. 5923--5930.

\bibitem{black2023zero}
K.~Black, M.~Nakamoto, P.~Atreya, H.~Walke, C.~Finn, A.~Kumar, and S.~Levine, ``Zero-shot robotic manipulation with pretrained image-editing diffusion models,'' \emph{arXiv preprint arXiv:2310.10639}, 2023.

\bibitem{kapelyukh2023dall}
I.~Kapelyukh, V.~Vosylius, and E.~Johns, ``Dall-e-bot: Introducing web-scale diffusion models to robotics,'' \emph{IEEE Robotics and Automation Letters}, vol.~8, no.~7, pp. 3956--3963, 2023.

\bibitem{mishra2024reorientdiff}
U.~A. Mishra and Y.~Chen, ``Reorientdiff: Diffusion model based reorientation for object manipulation,'' in \emph{2024 IEEE International Conference on Robotics and Automation (ICRA)}.\hskip 1em plus 0.5em minus 0.4em\relax IEEE, 2024, pp. 10\,867--10\,873.

\bibitem{mishra2023generative}
U.~A. Mishra, S.~Xue, Y.~Chen, and D.~Xu, ``Generative skill chaining: Long-horizon skill planning with diffusion models,'' in \emph{Conference on Robot Learning}.\hskip 1em plus 0.5em minus 0.4em\relax PMLR, 2023, pp. 2905--2925.

\bibitem{sridhar2024nomad}
A.~Sridhar, D.~Shah, C.~Glossop, and S.~Levine, ``Nomad: Goal masked diffusion policies for navigation and exploration,'' in \emph{2024 IEEE International Conference on Robotics and Automation (ICRA)}.\hskip 1em plus 0.5em minus 0.4em\relax IEEE, 2024, pp. 63--70.

\bibitem{wu2024tacdiffusionforcedomaindiffusionpolicy}
\BIBentryALTinterwordspacing
Y.~Wu, Z.~Chen, F.~Wu, L.~Chen, L.~Zhang, Z.~Bing, A.~Swikir, A.~Knoll, and S.~Haddadin, ``Tacdiffusion: Force-domain diffusion policy for precise tactile manipulation,'' 2024. [Online]. Available: \url{https://arxiv.org/abs/2409.11047}
\BIBentrySTDinterwordspacing

\bibitem{shuklaForceConditionedDiffusionPolicies2025}
R.~Shukla, S.~Moode, R.~Talan, N.~Dhanaraj, J.~H. Kang, and S.~K. Gupta, ``Force-{{Conditioned Diffusion Policies}} for {{Compliant Sheet Separation Tasks}} in {{Bimanual Robotic Cells}},'' in \emph{{{IEEE International Conference}} on {{Robotics}} and {{Automation}} ({{ICRA}})}, 2025.

\bibitem{shuklaLearningForceConditionedVisuomotor}
R.~Shukla, S.~Moode, R.~Talan, and S.~K. Gupta, ``Learning {{Force-Conditioned Visuomotor Diffusion Policy}} from {{Human Demonstrations}} for {{Complex Robotic Assembly Tasks}},'' in \emph{North {{American Manufacturing Research Conference}} ({{NAMRC}})}, 2025.

\bibitem{gao2022objectfolder}
R.~Gao, Z.~Si, Y.-Y. Chang, S.~Clarke, J.~Bohg, L.~Fei-Fei, W.~Yuan, and J.~Wu, ``Objectfolder 2.0: A multisensory object dataset for sim2real transfer,'' in \emph{Proceedings of the IEEE/CVF conference on computer vision and pattern recognition}, 2022, pp. 10\,598--10\,608.

\bibitem{gao2023objectfolder}
R.~Gao, Y.~Dou, H.~Li, T.~Agarwal, J.~Bohg, Y.~Li, L.~Fei-Fei, and J.~Wu, ``The objectfolder benchmark: Multisensory learning with neural and real objects,'' in \emph{Proceedings of the IEEE/CVF Conference on Computer Vision and Pattern Recognition}, 2023, pp. 17\,276--17\,286.

\bibitem{du2022play}
M.~Du, O.~Y. Lee, S.~Nair, and C.~Finn, ``Play it by ear: Learning skills amidst occlusion through audio-visual imitation learning,'' \emph{arXiv preprint arXiv:2205.14850}, 2022.

\bibitem{thankaraj2023that}
A.~Thankaraj and L.~Pinto, ``That sounds right: Auditory self-supervision for dynamic robot manipulation,'' in \emph{7th Annual Conference on Robot Learning}, 2023.

\bibitem{mejia2024hearingtouch}
J.~Mejia, V.~Dean, T.~Hellebrekers, and A.~Gupta, ``Hearing touch: Audio-visual pretraining for contact-rich manipulation,'' in \emph{2024 IEEE International Conference on Robotics and Automation (ICRA)}.\hskip 1em plus 0.5em minus 0.4em\relax IEEE, 2024.

\bibitem{lee2019making}
M.~A. Lee, Y.~Zhu, K.~Srinivasan, P.~Shah, S.~Savarese, L.~Fei-Fei, A.~Garg, and J.~Bohg, ``Making sense of vision and touch: Self-supervised learning of multimodal representations for contact-rich tasks,'' in \emph{2019 International conference on robotics and automation (ICRA)}.\hskip 1em plus 0.5em minus 0.4em\relax IEEE, 2019, pp. 8943--8950.

\bibitem{li2022seehearfeel}
H.~Li, Y.~Zhang, J.~Zhu, S.~Wang, M.~A. Lee, H.~Xu, E.~Adelson, L.~Fei-Fei, R.~Gao, and J.~Wu, ``See, hear, and feel: Smart sensory fusion for robotic manipulation,'' in \emph{CoRL}, 2022.

\bibitem{he2016deep}
K.~He, X.~Zhang, S.~Ren, and J.~Sun, ``Deep residual learning for image recognition,'' in \emph{Proceedings of the IEEE conference on computer vision and pattern recognition}, 2016, pp. 770--778.

\bibitem{perez2018film}
E.~Perez, F.~Strub, H.~De~Vries, V.~Dumoulin, and A.~Courville, ``Film: Visual reasoning with a general conditioning layer,'' in \emph{Proceedings of the AAAI conference on artificial intelligence}, vol.~32, no.~1, 2018.

\bibitem{ronneberger2015u}
O.~Ronneberger, P.~Fischer, and T.~Brox, ``U-net: Convolutional networks for biomedical image segmentation,'' in \emph{Medical image computing and computer-assisted intervention--MICCAI 2015: 18th international conference, Munich, Germany, October 5-9, 2015, proceedings, part III 18}.\hskip 1em plus 0.5em minus 0.4em\relax Springer, 2015, pp. 234--241.

\bibitem{ashish_attention_2017}
A.~Vaswani, N.~Shazeer, N.~Parmar, J.~Uszkoreit, L.~Jones, A.~N. Gomez, L.~Kaiser, and I.~Polosukhin, ``Attention is all you need,'' in \emph{Advances in neural information processing systems 30}, 2017.

\bibitem{Zhou_2019_CVPR}
Y.~Zhou, C.~Barnes, L.~Jingwan, Y.~Jimei, and L.~Hao, ``On the continuity of rotation representations in neural networks,'' in \emph{The IEEE Conference on Computer Vision and Pattern Recognition (CVPR)}, June 2019.

\end{thebibliography}

\end{document}